\documentclass[twocolumn]{article}

\usepackage{color}
\usepackage[cmex10]{amsmath}
\usepackage{array}
\usepackage[caption=false,font=footnotesize]{subfig}
\usepackage{url}
\usepackage{bm,graphics,graphicx}
\usepackage{algorithmic,multirow}
\usepackage{nicefrac}
\usepackage{float,natbib}

\begin{document}

\title{\textbf{Collective behavior recognition\\ using compact descriptors}}

\author{Gustavo F\"{u}hr\footnote{G. Fuhr is with Meerkat Vision LTDA} and Claudio R. Jung\footnote{C. Jung is with Institute of Informatics, Federal University of Rio Grande do Sul (UFRGS)}}
\date{Paper Submitted to Pattern Recognition Letters}


\maketitle

\vspace{2cm}

\begin{abstract}
This paper presents a novel hierarchical approach for collective behavior recognition based solely on ground-plane trajectories. In the first layer of our classifier, we introduce a novel feature called Personal Interaction Descriptor (PID), which combines the spatial distribution of a pair of pedestrians within a temporal window with a pyramidal 
representation of the relative speed to detect pairwise interactions. These interactions are then combined with higher level features related to the mean speed and shape formed by the pedestrians in the scene, generating a Collective Behavior Descriptor (CBD) that is used to identify collective behaviors in a second stage. In both layers, Random Forests were used as classifiers, since they allow features of different natures to be combined seamlessly.  Our experimental results indicate that the proposed method achieves results on par with state of the art techniques with a better balance of class errors. Moreover, we show that our method can generalize well across different camera setups through cross-dataset experiments.
\end{abstract}

\section{Introduction}

Automatic or semi-automatic analysis of human behavior has been studied by the computer vision community for several years. In particular, there has been increasing interest in inferring semantic information about the relation and interaction among people in a video sequence, denoted by group activity recognition, collective activity recognition or collective behavior recognition. In this paper, we chose the latter term because we are mainly interested in two aspects of the human behavior analysis: the activities that are being observed in the video and the social relation between pedestrians. 

Collective behavior involves two or more individuals interacting, and there are several ``levels of interaction''
that can be detected depending on the available information. If full 3D body information is known, one can detect
interactions such as caressing, hugging, holding hands, etc. If only position/orientation is available, events such
as walking together, chasing or approaching can be detected, while higher level information such as main flows can
be obtained in denser crowds for which individual pedestrians cannot be distinguished~\citep{Zhou2014}.

Although the literature on detection and classification of interactions among different people is rather vast,
this review focuses on techniques that explore pairwise and group interactions, which are related to the scope of this paper.

One of the first methods for collective behavior analysis was proposed by \citet{Oliver:2000}. 
They used Kalman filters to track the object locations, shape, color and velocity, and combine these data with the spatial relationship to nearby objects to describe interactions. For collective behavior, extensions to the well-known Hidden Markov Model (HMM) are presented to cope with multiple agents and state variables that interact with each other. Their algorithm can detect behaviors such as meet and continue together, meet and split and a person following another. 

\citet{Choi:2009} explore pedestrian tracking and human pose estimation (orientation) to define a set of descriptors 
called Spatial-Temporal Local (STL), which
encode the number and orientation of people within a radial area around an anchor subject across time. 
These features are fed to an SVM classifier that can detect events such as people queuing in line, talking and crossing a
street. Extracting the subjects orientations can be helpful to understand the role of an individual in an activity -- for instance, it 
is possible to distinguish waiting to cross a street or queuing by knowing the intended directions of the pedestrians. However, their approach
to obtain the orientations is highly dependent on the camera views used in the training and test sets.
An extension of the STL descriptor was presented by \citet{Chang:2015}, who use motion features together with the STL descriptor to learn pairwise relations for detecting different collective activities.

\citet{sun2016localizing} proposed a latent graph model to identify groups and their activities.
Their model explores pairwise intra- and inter-group properties through a hierarchical graph, being capable
of propagating multiple evidences among different layers of the graph. A graphical model was also explored by~\citet{Feng:2015},
focusing on the problem of social group identification, while fully connected conditional random fields that consider all
inter-agent interactions are explored in \citep{kaneko2014fully}.

\citet{Cheng:2014} represented the problem of group activity recognition using a three-layered approach that gathers information about the individuals performing the actions, the
possible pairs between two people and small groups.  First, each trajectory is modeled using
Gaussian processes. Additional information, such as location change since the
beginning of the sequence, the average velocity and the velocity ratio are also added as features.
 Context information is extracted by comparing both the location and velocity of an individual with relation to  others. They also propose the use of the shape information to describe an activity. These so-called action style features are
Histograms of Oriented Gradients (HOG) of the people in the group, which are shown
to discriminate correctly between different actions performed by individuals such as standing and fighting.
Similar to the work of \citet{Jacques:2007}, a geometric shape is used to analyze
the group formation. To achieve this goal, a Delaunay triangulation algorithm is applied
to the polygon that connects the people in the scene. Finally, these descriptors are used for training 
and classification: first, the descriptors are clustered using K-means to generate visual words, and then an 
SVM is used to classify the samples. 

\citet{group:detection:crowds:pami2015} tackled the problem of detecting social groups in crowds, leading to an intermediate representation between individuals and the whole crowd. They accomplish
this task by employing a correlation clustering method to a set of given trajectories, in which the affinity between crowd members is computed based on a set of specifically designed features characterizing both their physical and social identity. With the objective of detecting groups and their characteristics in a crowd scenario, \citet{shao2016learning} define priors that aim to add temporal smoothness and consistency for collective transitions -- groups are obtained by searching for pedestrians sets that fit well these priors.

\citet{lathuiliere2017recognition} presented an approach for group activity recognition
using single and two-person descriptors using Structured Support Vector Machines (SSVMs). For that purpose, they use a dictionary-based method, exploring geometric characteristics of the relative pose and motion between the two persons. 
\citet{kim2017discriminative} and \citep{wang2017recurrent} explored deep sequential models such as Long Short-Term Memory (LSTM) or Gated Recurrent Unit (GRU) in the context of behavior recognition. 
\citet{kim2017discriminative}  focused on discriminating
similar local motion patterns to detect group activities by proposing a discriminative group context feature  for sub-event detection, coupled with LSTMs or GRUs. \citep{wang2017recurrent} explored LSTMs to handle three level interactions, namely single human dynamics, within group human interaction and group to group interaction. Despite the good results achieved by these methods,
the authors had to use data augmentation to train their models due to the data hunger nature of their deep architecture.

This paper presents a new approach for joint pairwise and collective behavior detection using compact descriptors based solely on trajectory information, which are particularly suited when limited training data is available. We show that ground plane trajectories encode enough information to detect interaction and collective behavior with high accuracy and have the advantage of being robust to previously unseen scenarios, as indicated by the results in Section~\ref{sec:cross:dataset}. 
Our main contributions are the introduction of: 
i) compact yet discriminative descriptors for pairwise and collective interactions, and ii) a framework that uses pairwise interactions with high semantic value to recognize collective behavior. The proposed approach is presented next.

\section{The Proposed Method}
\label{sec:proposed}

The input of the proposed approach is a set of ground plane positions from the pedestrians. A self-calibration
method that explores pedestrian detection/tracking (e.g., \citep{Fuhr:2015}, which is used in this work) can be used
to map from image to world coordinates, and pedestrian tracking methods tuned for calibrated cameras 
can be used~\citep{fuhr2014combining} to generate trajectories 
(see~\citep{smeulders2013visual} for a survey on tracking).

\begin{figure}[ht!]
  \centering
     \vspace{-.2cm}
  \includegraphics[width=.5\textwidth]{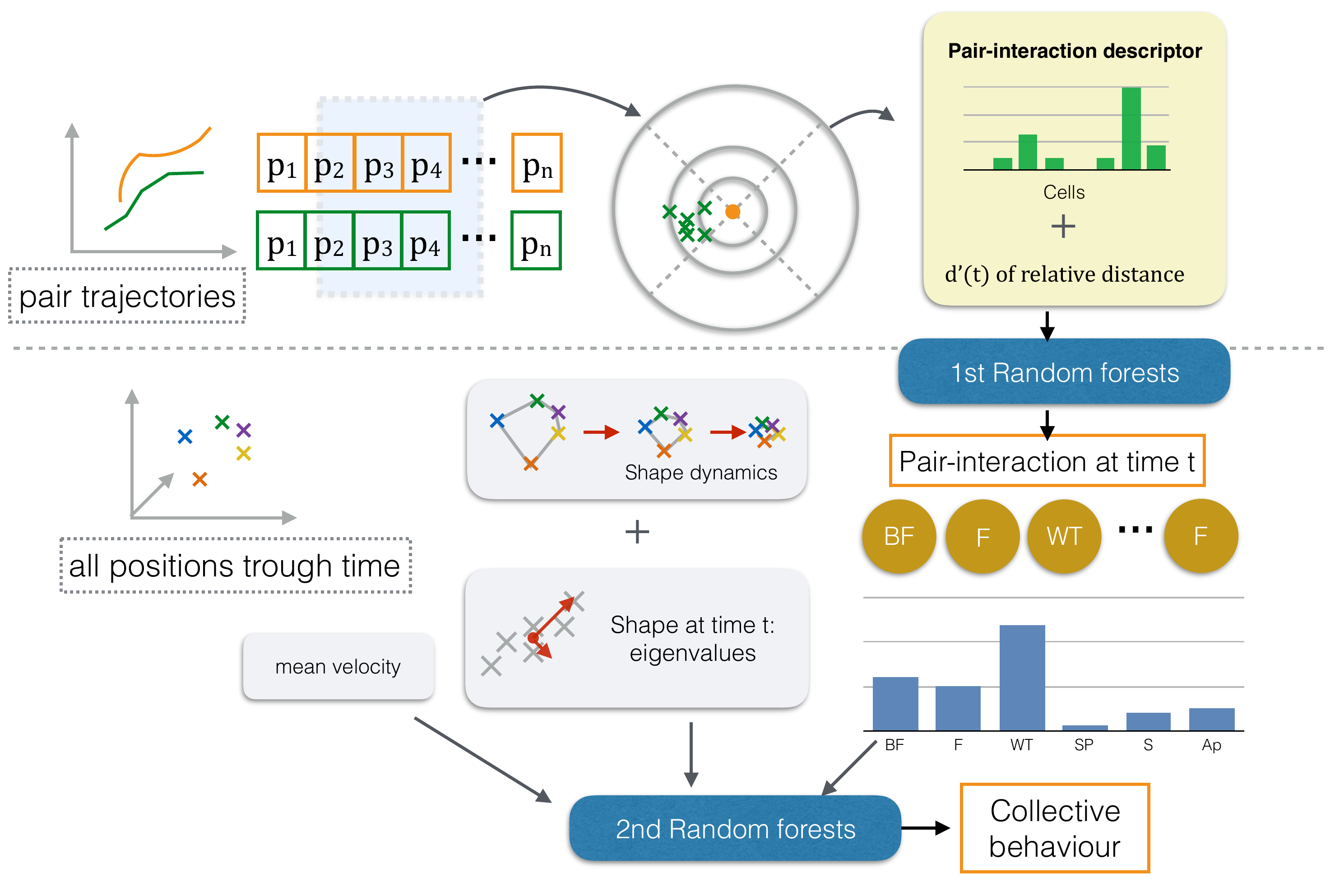}
  \caption{ \label{fig:method_schema}
    \vspace{-.5cm}
 Overview of the proposed method.}
     \vspace{-.4cm}
\end{figure}

Given a pair of trajectories related to two people, we proposed a Personal Interaction Descriptor (PID) to encode their
relationship and classify it using a Random Forest, since it allows features of different natures to be combined seamlessly~\citep{Breiman01randomforests} and have shown highly accurate results against more commonly used classifiers~\citep{fernandez2014we}.
We then group pairwise interactions into a Collective Behavior Descriptor (CBD), which is fed with group shape information to a second
classification layer to identify collective behavior. A schematic overview of the method is illustrated in Fig.~\ref{fig:method_schema}.

\subsection{Pairwise interactions}
\label{sec:interactions}

Let a unidirectional interaction between two pedestrians $o_1$ and $o_2$ be denoted by $o_1 \rightarrow o_2$, where $o_1$ is the anchor subject, and $o_2$ is the target subject. 
Although there are several possible interactions, we defined in this work 
six different pairwise interactions inspired on ~\citep{Choi:2009}: 
\emph{being-followed} (BF), \emph{following} (F), \emph{walking-together} (WT), 
\emph{standing pair} (SP), \emph{splitting} (S), and \emph{approaching} (Ap). In fact, the main difference from~\citep{Choi:2009}
is that their ``follow'' behavior was split into 
following and being-followed here, recalling that our interactions are not symmetric in general.

\begin{figure}[ht!]
  \centering
    \subfloat[]{\label{fig:pid:radius}\includegraphics[height=3.2cm]{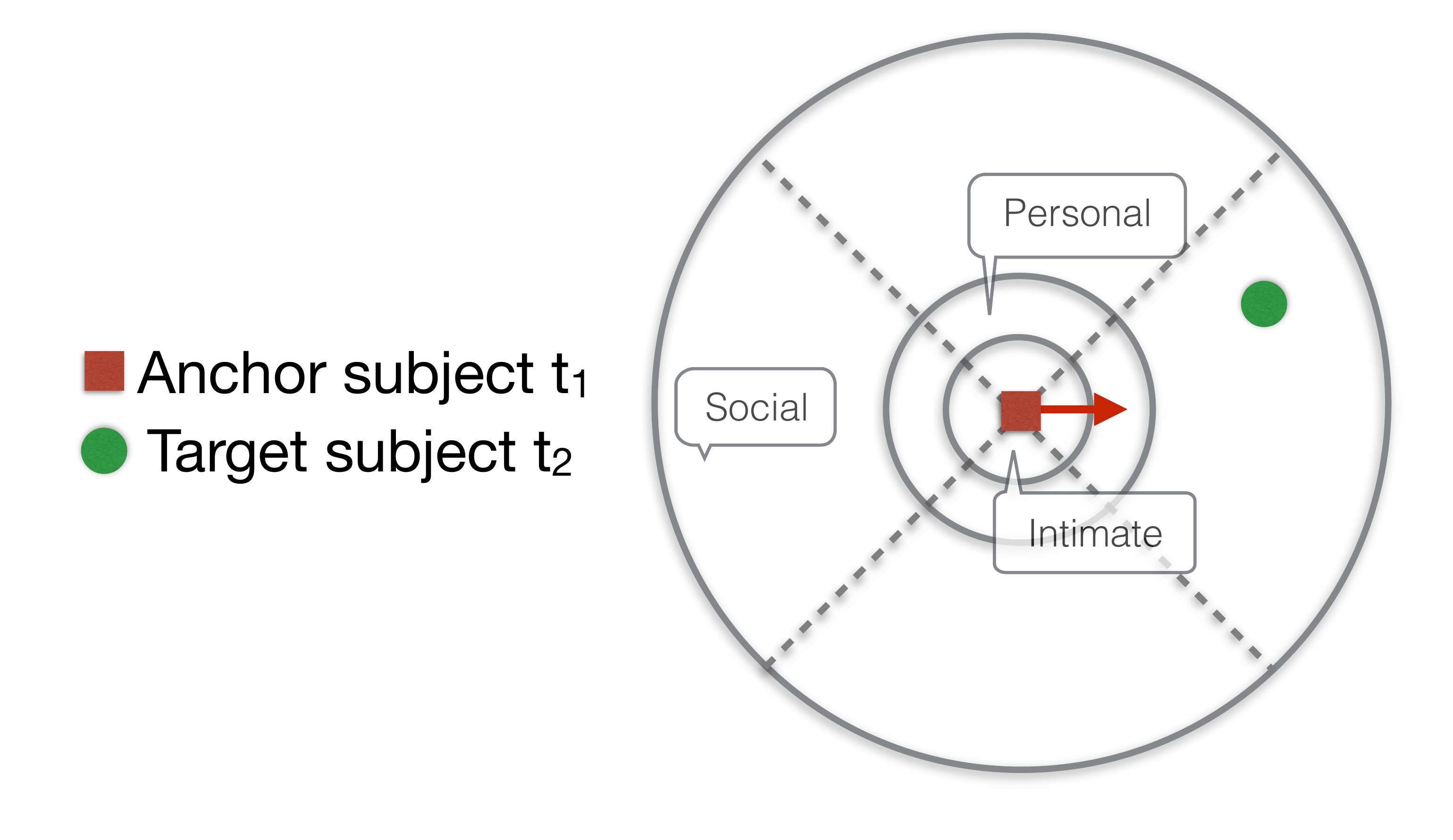}}
  \subfloat[]{\label{fig:pid:gaussian}\includegraphics[height=3.2cm]{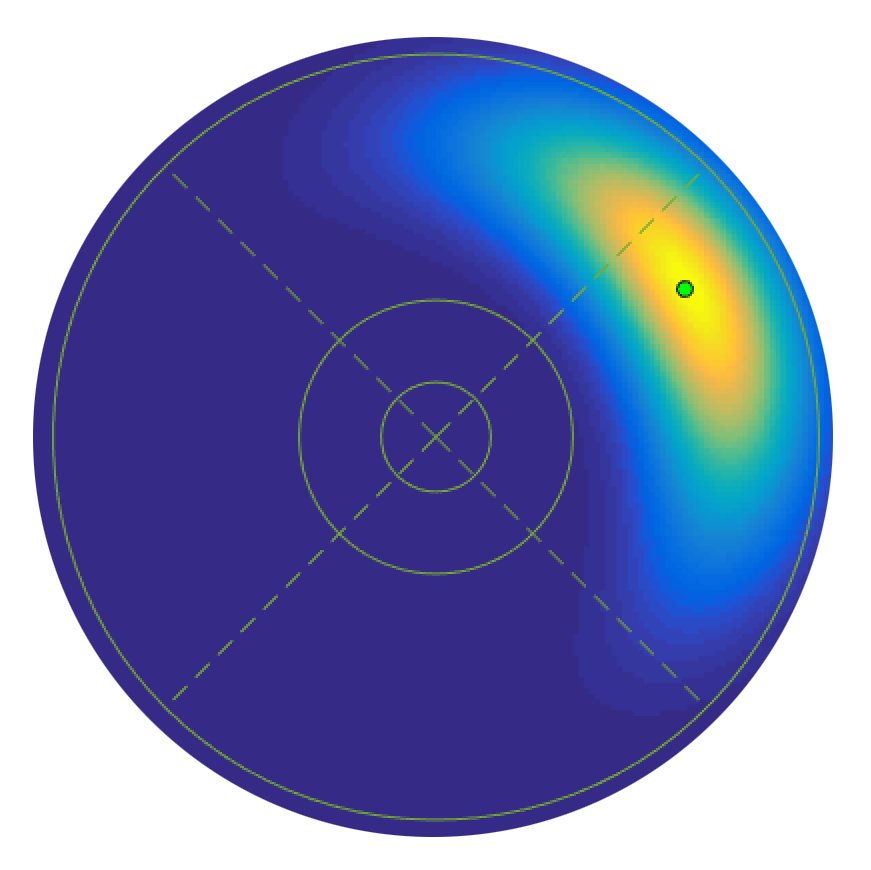}} 
  \caption{(a) Four angles and three distances are used to divide the region around a subject into bins/cells. (b) A normal
  distribution is used to introduce a soft boundary between cells.}
  \label{fig:pid}
\end{figure}

The first cue that is important for identifying pairwise interactions is the set of relative positions between two subjects within a time window, which can be explored to detect if they are near/far each other, side by side and so on. 
In this work, we divide the region around a subject into cells using boundaries in the polar domain (both distances and angles) w.r.t. the anchor subject. One descriptor is created for each of the neighboring subjects, i.e., we define a pairwise descriptor. More precisely, the  distances boundaries
in our Personal Interaction Descriptors (PIDs) 
were obtained according to the studies of \citet{Hall:1973}, which sets different radii for the expected intimate (up to 0.5 meters), personal (0.5 to 1.25 meters), social (1.25 to 3.5 meters) and public (more than 3.5 meters) interactions between two subjects.  In fact, the concept of proxemics was already used in the context of group detection in methods such as~\citep{Jacques:2007,group:detection:crowds:pami2015}.
Additionally, we divide the region into four equidistant angular sectors to identify back, front
left and right sides of the anchor subject. The cell disposition around the anchor is illustrated in Fig.~\ref{fig:pid:radius}. 

It is worth noticing that the spatial distribution of the neighboring pedestrians was also explored in~\citep{Choi:2009,Choi:2014}.
However, they used a classifier to obtain the orientation of each one of the anchor's neighbors, which is based on image features and might be sensitive to the camera setup,
and that leads to a higher-dimensional representation. 
Instead of using image features, we estimate the local orientation of a moving pedestrian based on the corresponding trajectory, filtered by a Double Exponential Smoothing technique~\citep{Laviola:2003} to reduce the effect of trajectory jitter.
However, one limitation of the proposed method is that it fails for stationary pedestrians, or for slowly moving ones
(for which the orientation estimation is very noisy).  In fact, when the speed of the anchor subject is smaller
than a threshold $T_s$, we disregard the orientation part of the proposed descriptor by setting a random orientation
value for the target pedestrian, so that there is no bias to a particular orientation (and the information is encoded by the distance only).

We then compute a histogram of relative positions between the targets within
a temporal window using the spatial bins illustrated in Fig.~\ref{fig:pid:radius}. More precisely,
for a given frame $f_t$ at time $t$, we analyze a temporal window with $T_1$ frames (assumed to be a power of two) centered at $t$, i.e. 
from $f_{t-T_1/2+1}$ to $f_{t+T_1/2}$. A traditional histogram
is obtained by simply counting the number of relative positions that lies in each bin along all frames in the window. However, this approach
is very sensitive to samples that lie on close to the boundaries between bins, so that similar pairwise behaviors
may generate considerably  different histograms. This problem is amplified when the number of samples is small, as it is often the case on human behavior classification. To alleviate this problem, we adopt a soft assignment approach based on histogram is kernel density estimation (KDE)~\citep{kde_comparison_tsp94}, in which 
a kernel centered at each observation is used to obtain a continuous PDF of the data. In this work, we use a Gaussian kernel defined in polar
coordinates $(\rho, \theta)$ given by (disregarding normalization):
\begin{equation}
G_{\text{KDE}}(\rho, \theta; \rho_{o}, \theta_{o}) =  \exp\left[ -{\frac{(\rho - \rho_{o})^2}{2\sigma_{\rho}}}  -{\frac{d_{\theta}(\theta, \theta_{o})^2}{2\sigma_{\theta}}}\right ],
\label{eq:gaussian}
\end{equation}
where $\rho_{o}$ and $\theta_{o}$ are the polar coordinates of the target subject relative to the anchor (i.e. the location of the sample), $d_{\theta}$ is a function that computes the smallest difference between two angles, and $\sigma_{\rho}$, $\sigma_{\theta}$
are the scale parameters in the distance and orientation domains, respectively.

To obtain the histogram, one could just integrate the kernel-smoothed PDF over each bin. In this work, 
such integral is approximated by sampling a constant number of points around its center $(\rho_{o}, \theta_{o})$ 
equally spaced in a $K_s\sigma_{\rho} \times K_s\sigma_{\theta}$ grid,
normalizing and then summing over each spatial cell, where $K_s$ controls the extent of the sampling region. 
If $h_{o_t}(c)$ denotes\footnote{For the following expressions in this section, let the target subject $o_2$ at time $t$ be expressed as $o_t$.} the histogram for target $o$ at time $t$ at cell $c$, it is given by

\begin{equation}
h_{o_t}(c) = \sum_{\tau=t-\frac{T_1}{2}+1}^{t+\frac{T_1}{2}} a_g(\tau) \sum_{\bm{s} \in S_{o_{\tau}}} \chi_c(\bm{s}) G_{\text{KDE}}(\bm{s}; \bm{p}_{o_t}  )
\label{eq:histogram},
\end{equation}
where
\begin{equation*}
\chi_c(\bm{x}) = \begin{cases}
1 &\text{~if~} \bm{x} \in A_c,\\
0 &\text{otherwise}.
\end{cases}
\end{equation*}
is the indicator function for cell $c$ ($A_c$ is the spatial region that defines $c$),
$S_{o_t}$ is the set of samples from the grid generated by target $o_t$, $\bm{p}_{o_t}$ contains the polar coordinates of $o_t$, $a_g(t)$ is a normalization factor
for the samples generated at frame $t$ such that the final histogram sum is equal to one. 
An illustration of soft-assignment induced by the KDE is provided in 
Fig.~\ref{fig:pid:gaussian}. 

The second cue in  PID is the dynamic aspect of a pair trajectory. The KDE-smoothed histogram
encodes the cumulative relative position of the target agent w.r.t. the anchor, but temporal information is lost.
For instance, in the approaching and splitting interactions, the distances decrease and increase, respectively, in a reasonably-sized temporal window. However, such information is lost when computing the histogram. 

To include this information in the descriptor, we compute the relative speed $d'(t)$, where $d(t)$ is
the relative distance between the pedestrians under analysis. 
We evaluate $d'(t)$ within the temporal window in a pyramidal fashion, and append
these values to the PID. At the first level of the pyramid, we take the mean of the derivatives for the whole temporal window. Next, we evaluate
the averages at the first and second half, generating two values and so on. More precisely, 
for each level $l \in \{0, 1, \dots, l_{max}\}$, where $l_{max}$ is 
the highest level, we build a $2^l$-dimensional feature vector 
\begin{equation}
\bm{d}'_l(t) = (\mu^l_1(t),\cdots,\mu^l_{2^l}(t))^T,
\label{eq:dprime}
\end{equation} where
\begin{equation}
\mu^l_k(t) = \frac{2^l}{T_1}\sum_{\tau = \frac{T_1}{2^l}(k-1)}^{\frac{T_1}{2^l}k-1}d'\left(t+\tau-\frac{T_1}{2}+1 \right)
\label{eq:mu}
\end{equation} 
is the average of $d'(t)$ in the $k^{th}$ partition interval at level $l$. Finally, a consolidated
vector $\bm{d}'(t)$ is obtained by concatenating $\bm{d}'_l(t)$ for $l=0,...,l_{max}$, and it encodes
hierarchical information on the relative speed for the pair of pedestrians within the analyzed time window. The dimensionality of $\bm{d}'(t)$ is $2^{l_{max}+1}-1$, which is much smaller than using different time intervals to compute the spatial histograms, as proposed in~\citep{Choi:2009}.

\subsection{Collective behavior recognition}
\label{sec:collective}

The final goal of our method is to use all the detected pairwise interactions and to classify the collective activity of a given group. Throughout this paper, our main hypothesis is that there are different cues of information that are required to describe an interaction or activity. Not surprisingly, our collective descriptor also uses Random Forests to mix different kinds of information. Once more, we extract data  in a given temporal window of $T_2$ (also a power of two) frames and assume that a single activity appears in the scene (as in \citep{Choi:2014,Amer:2014}), and eventual additional trajectories are rejected in our experiments. Despite the fact that our collective method provides a general approach for inferring group activities, in this paper we chose to study a subset of them that commonly appear in surveillance systems and represent a good sample of the activities seen in real-world applications: \emph{Gathering}, \emph{Talking}, \emph{Dismissal}, \emph{Walking}, \emph{Chasing} and \emph{Queuing}. These interactions were also tackled in~\citep{Choi:2012}, which also
allows comparisons with other methods.

 Different types of pairwise interactions are expected to arise in a single collective activity. 
For instance,  let us consider the queuing event. 
People waiting in line are related trough a \emph{standing-pair} interaction  and, if a  person is directing him/herself to the line, an \emph{approaching} interaction is observed. Even more,
 if two or more people are advancing significantly in that line, our definition would indicate that there are \emph{following} and \emph{being-followed} interactions appearing. 
 
To describe the multitude of such interactions, our Collective Behavior Descriptor (CBD) starts by building a histogram of the pairwise interactions detected by our first classifier within the temporal window. 
This histogram is normalized to account for variations in the number of people that compose the group, such as its sum is always equal to one. Since we defined six interactions 
 in Section~\ref{sec:interactions}, this histogram represents the first six dimensions in our descriptor.

However, only using pairwise interactions is not enough to characterized the collective behavior of a group. It is also important to use
features of the group itself, formed by all pedestrians in the group. In particular, the mean speed of the group can be useful do distinguish
static from dynamic groups. In this work, we define the mean speed $v_{\mu}(t)$ of the group at each frame $t$ as an average of the
individual speeds within the temporal window:
\begin{equation}
v_{\mu}(t) = \frac{1}{T_2}\sum_{\tau=t-\frac{T_2}{2}+1}^{t+\frac{T_2}{2}} \frac{1}{\#S_{\tau}} \displaystyle\sum_{\bm{s} \in S_{\tau} } \| \bm{v}_{s\tau} \|,
\label{eq:velocity}
\end{equation}
where $S_{\tau}$ is the set of subjects in the group at time $\tau$ and $\bm{v}_{s\tau}$ is the velocity vector of pedestrian
$s$ at frame $\tau$. 

Another relevant source of information is the spatial distribution of the pedestrians along the group, and how it
changes in time.  For instance, in behaviors such as gathering and dismissal, the group goes from disperse to
compact and from compact to disperse, respectively. To encode the temporal variation of the group dispersion, we first
define the group dispersion $\delta(t)$ at frame $t$ as
\begin{equation}
\delta(t) = \sqrt{ \frac{1}{\#S_{t}} \sum_{ s \in S_{t} }\left\| \bm{p}_{st}-\bm{\mu}_t \right\|^2},
\end{equation}
where $\bm{p}_{st}$ is the position of a subject $s$ and $\bm{\mu}_{t}$ is centroid of the group
at frame $t$. Finally, the temporal dispersion change $s(t)$ is given by
\begin{equation}
s(t) = \frac{1}{T_2}\sum_{\tau=t-\frac{T_2}{2}+1}^{t+\frac{T_2}{2}} \delta'({\tau}),
\label{eq:variance}
\end{equation}
i.e. it is the average of the derivatives along the temporal window. 
For both $s(t)$ and $v_{\mu}(t)$, we also experimented with the same pyramidal averaging approach used for the derivatives described in Section~\ref{sec:interactions}, yet the overall classification performance remained almost constant. Therefore, we decided to use just the temporal
average (first level of the pyramid) to keep the descriptor compact.

\begin{figure}[ht!]
  \centering
  \subfloat[Gathering]{\includegraphics[width=0.24\textwidth]{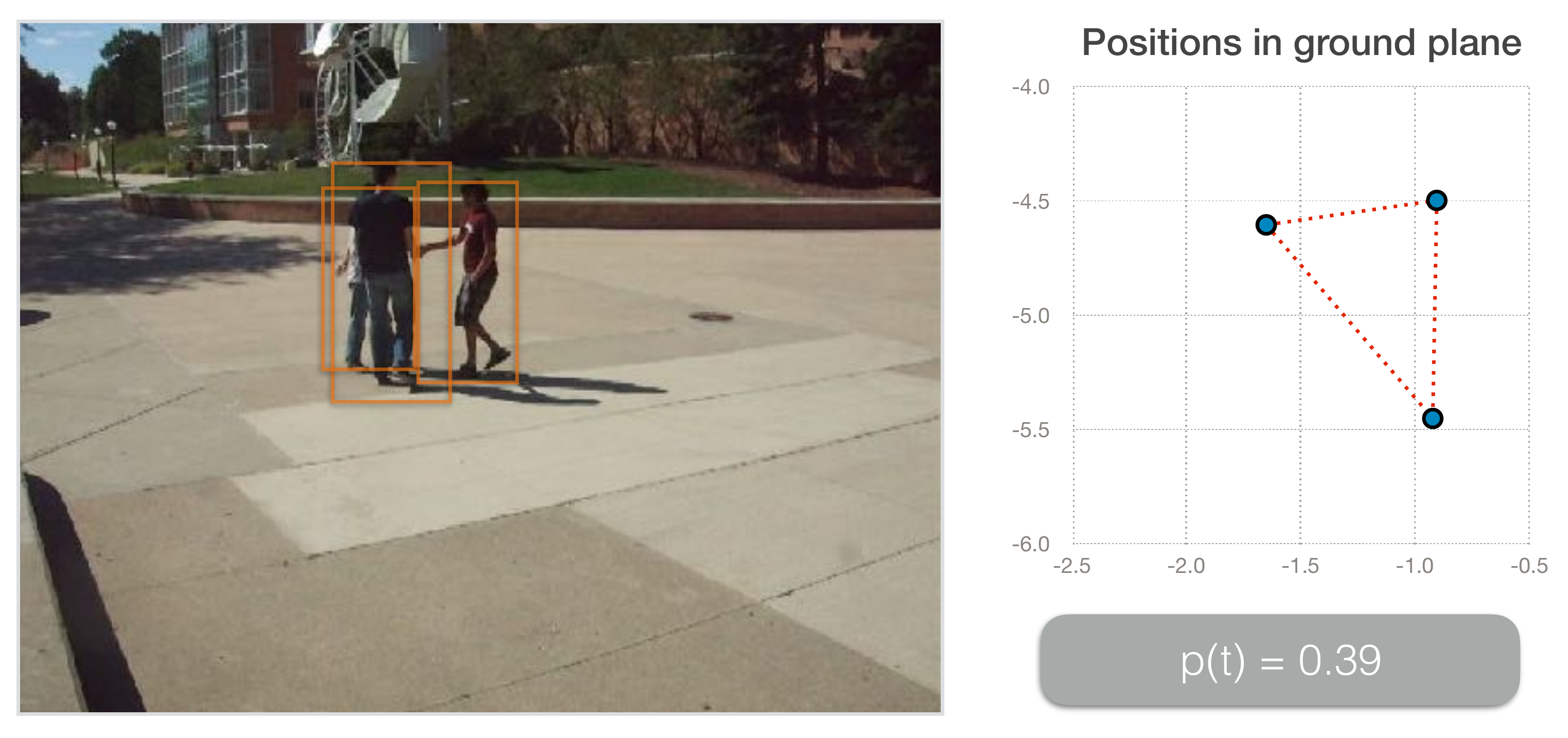}} 
  \subfloat[Queuing]{\includegraphics[width=0.24\textwidth]{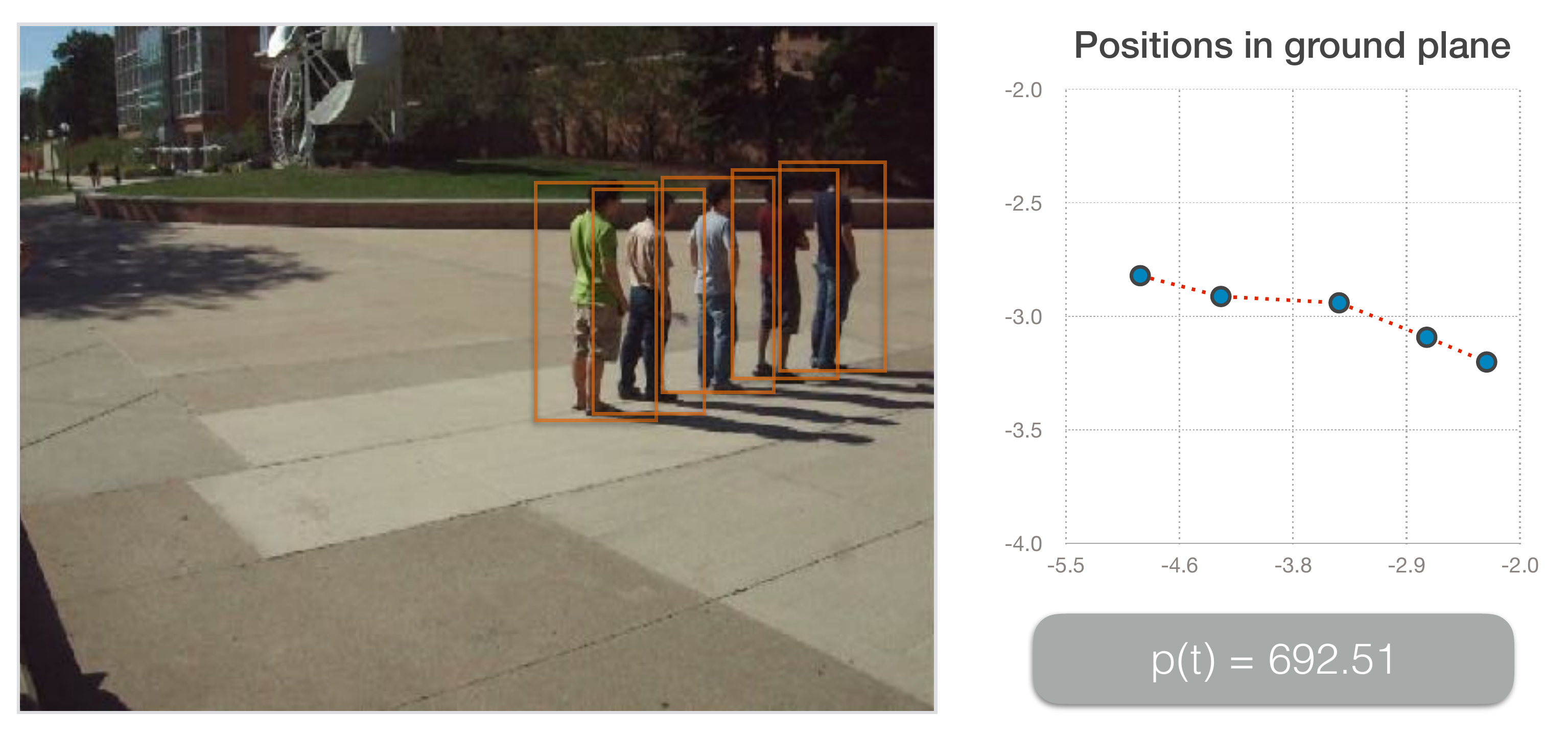}}
  \caption{The value of $p(t)$ for two different activities.}
  \label{fig:pca_groundplane}
\end{figure}

The final information cur used in CBD aims to capture information about the shape of the group. The dispersion feature $\delta(t)$ indicates
how close to each other the members of the group are, but does not capture the shape (or the orientations in which the dispersion occurs). 
In particular, the alignment of the group members provide cues about the the nature of the group: people in a conversation are typically arranged in a polygonal-like formation, opposed to queuing situations in which people are roughly aligned. Hence, we also use a group descriptor $p(t)$ given by
\begin{equation}
p(t) = \begin{cases}
\nicefrac{\lambda_{max}}{\lambda_{min}}  &\text{if $\#S_g > 1$ and $\lambda_{min} > 0$}\\
0 &\text{otherwise}
\end{cases},
\label{eq:pca}
\end{equation}
where $\lambda_{min} \leq \lambda_{max}$ are the two non-negative eigenvalues of the covariance matrix of a set of 2D points
that represent the subjects ground plane positions at time $t$. When pedestrians are aligned, $\lambda_{min} << \lambda_{max}$
and $p(t)$ is larger; on the other hand, smaller values for $p(t)$ are expected in a more regular distribution of pedestrians
within the group, as illustrated in Fig.~\ref{fig:pca_groundplane}.

\section{Experimental Results}
\label{sec:experiments}

This section aims to evaluate the accuracy of the proposed interaction and collective behavior classifiers. 
Additionally, we would like to assess which cues of information are indeed relevant for inferring interactions
and collective activities. Since we used Random Forests for the two levels of inference (pairwise interactions and
collective activity), each feature in our descriptors is treated individually in the classification phase. For a different classification method, such as SVMs, it would be tricky to verify the importance of each feature dimension. For instance, if one or more values related to a given cue are added to the descriptor, it would be difficult to see if either the distance function, the feature normalization or the information itself is to blame for a possible decrease in classification accuracy. On the other hand, since Random Forests classifiers take the features ``as they are'' and treat them independently, there is a more direct relation between cue importance and performance, which enables us to see clearly which components of our systems are responsible for the overall performance of the method.

As in recent papers for collective behavior detection, we used the \textit{New Collective Activity Dataset} (N-CAD) proposed in~\citep{Choi:2012}, which has been gaining a lot of popularity
in the last years and is publicly available in the author's website\footnote{\url{http://www-personal.umich.edu/~wgchoi/eccv12/wongun_eccv12.html}.}. 
The dataset is composed of 33 different sequences with 6 different collective behaviors: 
\emph{Gathering}, \emph{Talking}, \emph{Dismissal}, \emph{Walking}, \emph{Chasing} and
\emph{Queuing}. We carefully annotated all the pairwise interactions that appear in the sequences using the six interaction described in Section~\ref{sec:interactions}, making them available together with our source code\footnote{\url{https://github.com/gustavofuhr/collective_behavior}}. 

All experiments in the N-CAD dataset were performed using the same 3-fold validation scheme proposed by \citet{Choi:2012}.
We used $T_1=T_2=64$, which showed a good compromise between robustness to noise and detection lag.

\subsection{Analysis of the Interaction Descriptor}

\begin{figure}[ht!]
  \centering
  \subfloat[]{\label{fig:int:derivatives}\includegraphics[width=0.27\textwidth]{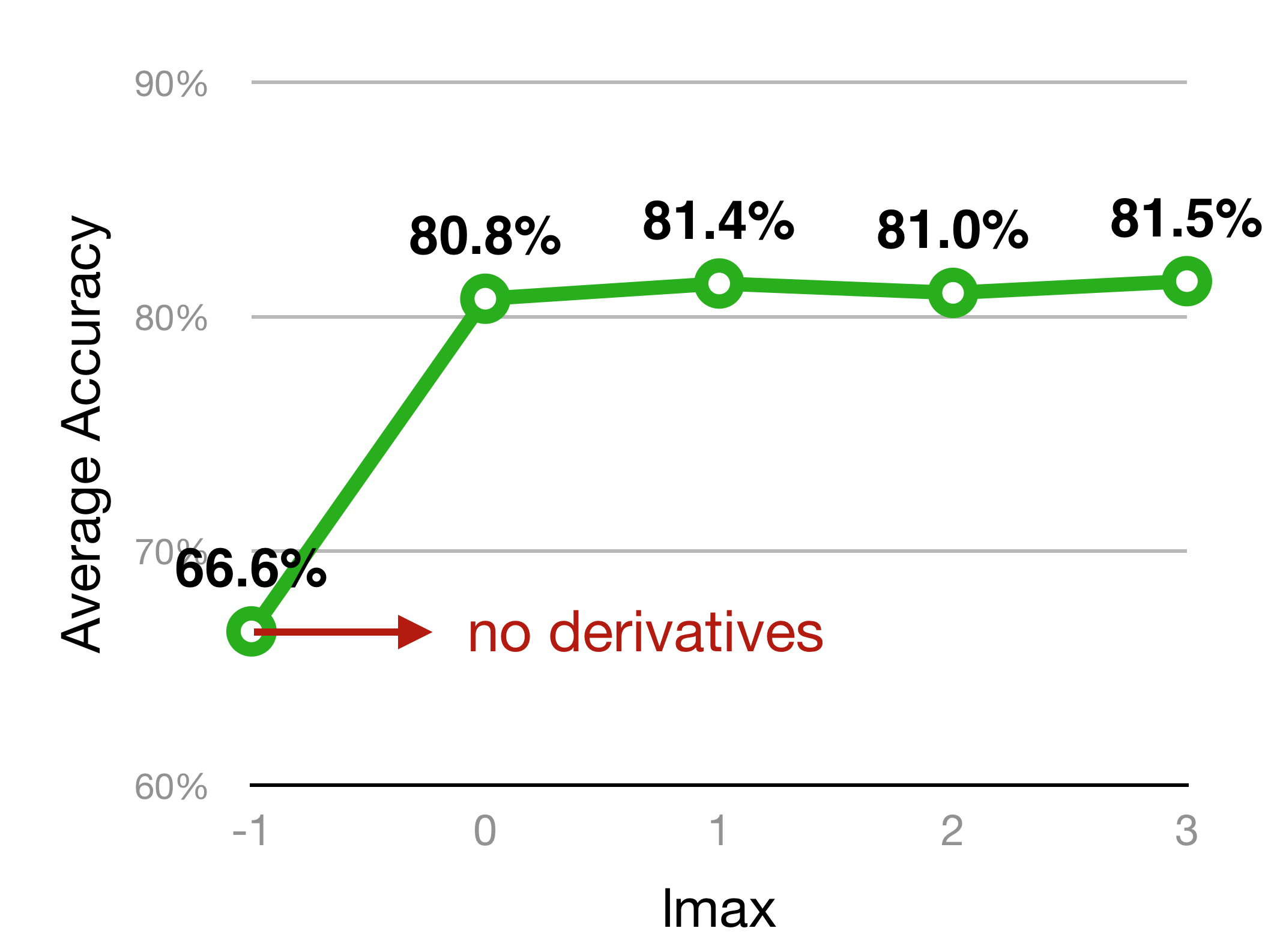}} 
  \subfloat[]{\label{fig:int:cmat}\includegraphics[trim={1cm 5cm 1cm 5cm}, clip, width=0.22\textwidth]{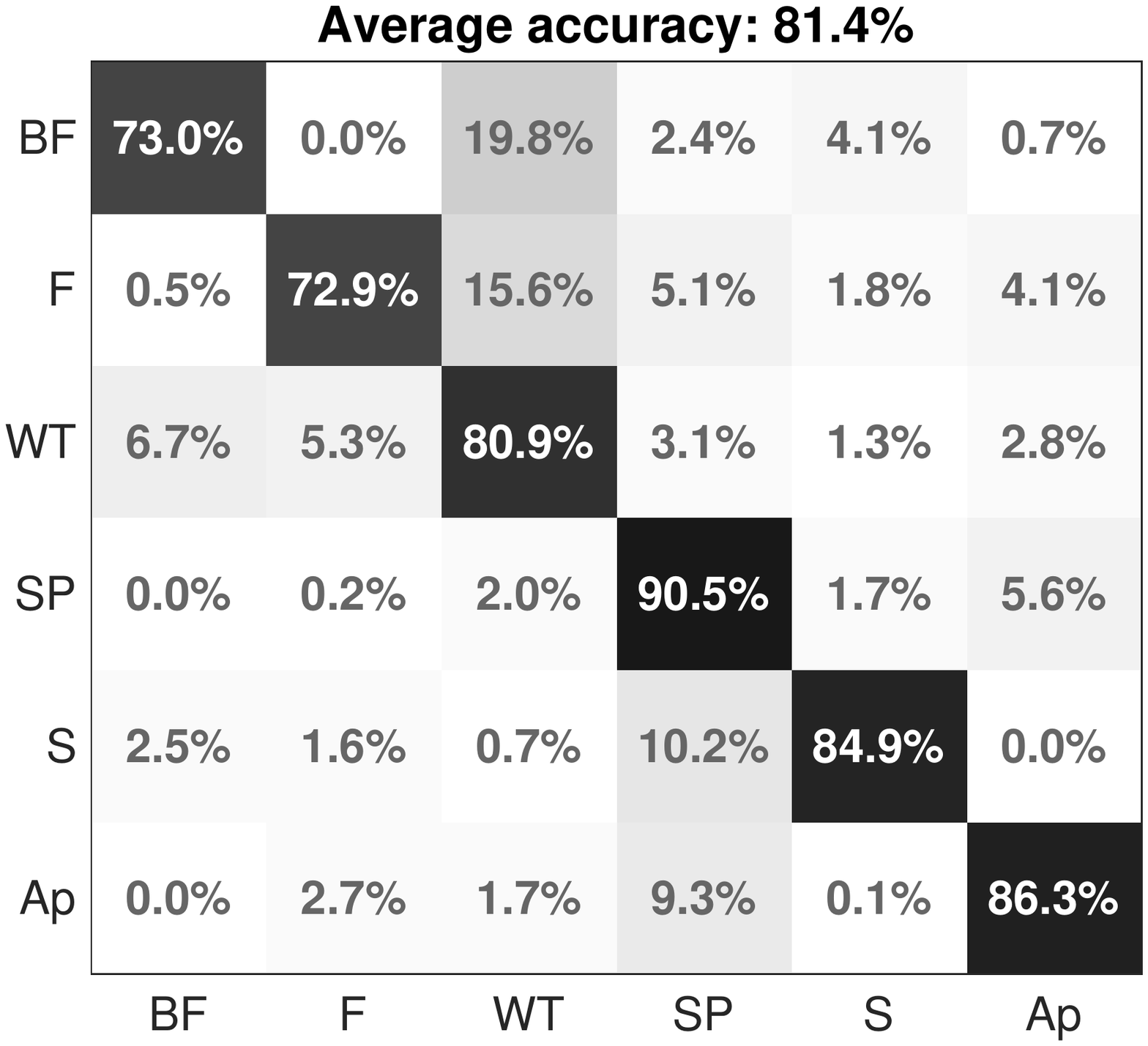}} 
  \caption{Experimental results for our interaction descriptor. (a) Impact of the number of levels in the pyramidal representation.
  (b) Confusion matrix for $l_{max}=1$ (chosen default value). }
  \label{fig:int}
\end{figure}

The first set of experiments aims to attest if our interaction descriptor (PID) is indeed discriminative despite its very small size (usually less than 20 dimensions). For these experiments we use a step of $5$ frames in classification/testing and we defined our standard deviation for the PID at $\sigma_{\theta} = \pi/8$ (so that each sector in the histogram contains four standard deviations in the orientation) and $\sigma_{\rho} = 0.25$, based on 
experiments. We used $K_s = 3$ when generating samples for the KDE, which covers virtually all the area under the Gaussian kernel. 
To define the number of levels $l_{max}$ used in the speed pyramid, we computed the accuracy for different values as shown in Fig.~\ref{fig:int:derivatives}.
Based on this plot, we believe that $l_{max} = 1$ presents a good compromise between descriptor size and accuracy, and this value
was chosen as the default.

Fig.~\ref{fig:int:cmat} depicts the confusion matrix of our pairwise interaction detector, with an average per-class accuracy of 84.3\% and minimum individual accuracy of 78\% for \emph{following} and \emph{being-followed}. 
It is worth noticing that \emph{following} and \emph{being-followed} sometimes are classified as \emph{walking-together}, around 16\% to 20\% of times. This is mostly due to noise in the estimates for the ground plane position. For example, when a group of several people is walking in two or more rows, we annotated that the subjects in the back rows are following the ones in front of them, even if the whole group is walking as one. When this group is far from the camera, mapping from image to world coordinates gets more sensitive to noisy observations, and the relative positions in our histograms can present jitter. As a consequence, there might be a mix up between the front (or back) to bins in the subject to the left or right sides. 

Finally, we ran the same experiments using hard boundary assignments to compute the histograms required in the PID (instead
of KDE). Results were in average 1.5\% lower, indicating that KDE-smoothing is indeed useful.

\subsection{Collective Behavior Recognition}

The second set of experiments evaluated the accuracy of the collective behavior recognition method proposed in this paper, using the same 3-fold cross-validation scheme. To evaluate the effect of different cues, we tested three different versions of our approach by progressively adding the cues. More precisely, we started by using only interaction histograms and pyramidal mean velocities, then added the spatial distribution dynamics encoded in Eq.~\eqref{eq:variance}, and finally added the shape cue encoded from  the eigenvalue ratios. The average accuracy obtained in this experiment were 75.5\%, 78.2\% and 87.2\%, indicating that the full set of cues leads to the best result.

The per-class accuracy of our method and competitive approaches,
as well as the mean per-class accuracy (MPCA) and the standard deviation, 
are reported in Table~\ref{tab:ncad}. It is interesting to note that our lowest per-class accuracy was 
$81\%$, which is the highest among all compared methods. Furthermore, our approac presents a homogeneous error distribution among the classes, as indicated by the lowest standard deviation.
Our overall accuracy was similar to~\citep{sun2016localizing} and~\citep{Amer:2014}, and better than recent approaches such as SSVM~\citep{lathuiliere2017recognition} and RMIC~\citep{wang2017recurrent}.
 It should be noticed that the best overall method DCGF+GRU~\citep{kim2017discriminative} was trained and evaluated with augmented data, so that a direct comparison of the results might be biased.

\begin{table}
\centering
\caption{Comparison with state-of-the-art methods for the N-CAD dataset}
\resizebox{\columnwidth}{!}{
\begin{tabular}{lcccccccc}
\hline
\multirow{2}{*}{\textbf{Method}} & \multicolumn{8}{c}{\textbf{Activity}} \\
\cline{2-9} 
& Chasing & Dismissal & Gathering & Queueing & Talking & Walking & MPCA & Std\\
\hline
Our (PID+CBD) & 84.0 & 81.0 & \textbf{90.0} & 86.0 & 90.0 & 92.0 & 87.2 & \textbf{4.2}\\
LRG~\citep{sun2016localizing} & \textbf{99.2} & 84.3 & 55.8 & 99.5 & 91.5 & 94.8 & 87.5 & 16.5 \\
SSVM~\citep{lathuiliere2017recognition} & 96.9 & 69.1 & 57.1 & 89.9 & 70.5 & \textbf{97.6} & 80.2 & 16.9\\
HiRF~\citep{Amer:2014} & 98.2 & 87.6 & 54.9 &   99.2 &   89.3 &   94.3 & 87.3 & 16.5 \\
HiGM~\citep{Choi:2012} &  91.9 &   77.0 &  43.5 &   93.4 &   82.2 &   87.4 & 79.2 & 18.5 \\
\citet{Chang:2015} & 53.9 & \textbf{90.5} & 59.9 & 86.3 & \textbf{97.0} & 94.3 & 80.3 & 18.6 \\
DCGF+GRU~\citep{kim2017discriminative} &  96.1 &   87.3 &   79.6 &  \textbf{100.0} &  90.6 &   94.0 &   \textbf{91.3} & 7.2 \\
RMIC~\citep{wang2017recurrent} &  89.2 & 68.4 &  71.9 & 95.5 & 86.6 &   95.9 &  84.6 & 11.8
\label{tab:ncad}
\end{tabular}}
\end{table}

\begin{figure}[ht!]
  \centering
\subfloat[Chasing]{\label{fig:collective_frames:cha}\includegraphics[height=0.11\textwidth]{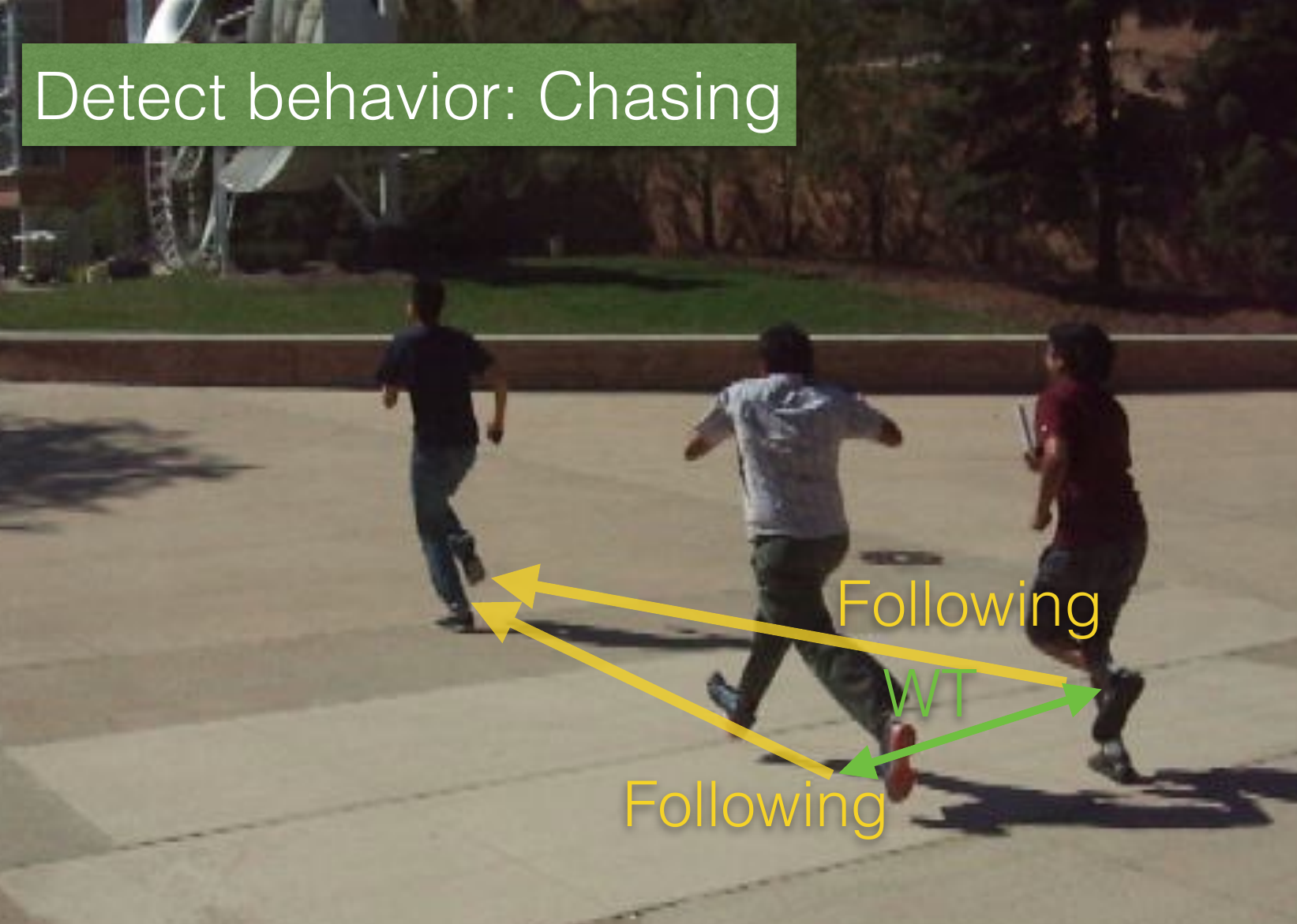}}~
\subfloat[Gathering]{\label{fig:collective_frames:gat}\includegraphics[height=0.11\textwidth]{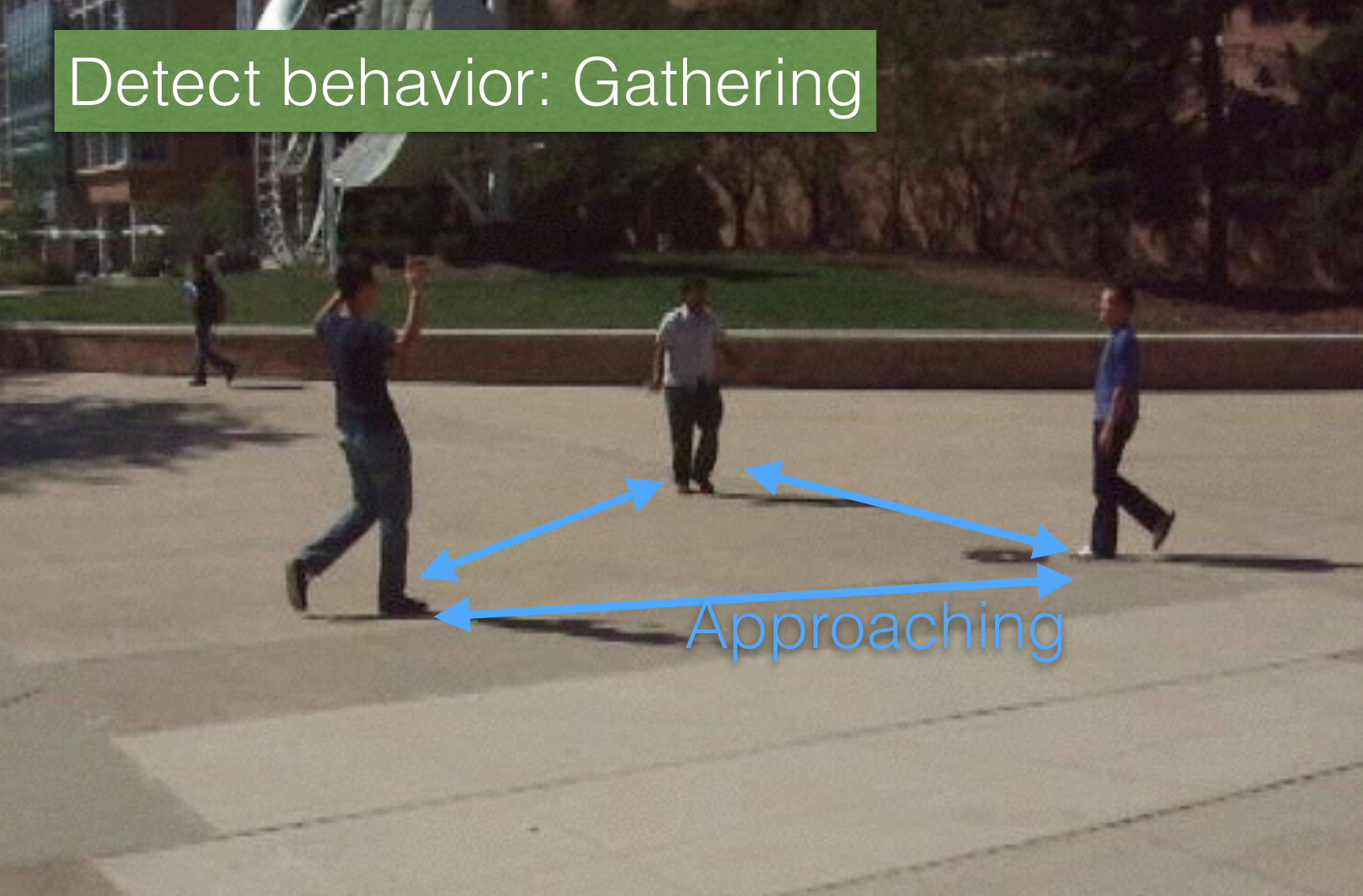}}~ 
\subfloat[Queueing]{\label{fig:collective:full:queueing}\includegraphics[height=0.11\textwidth]{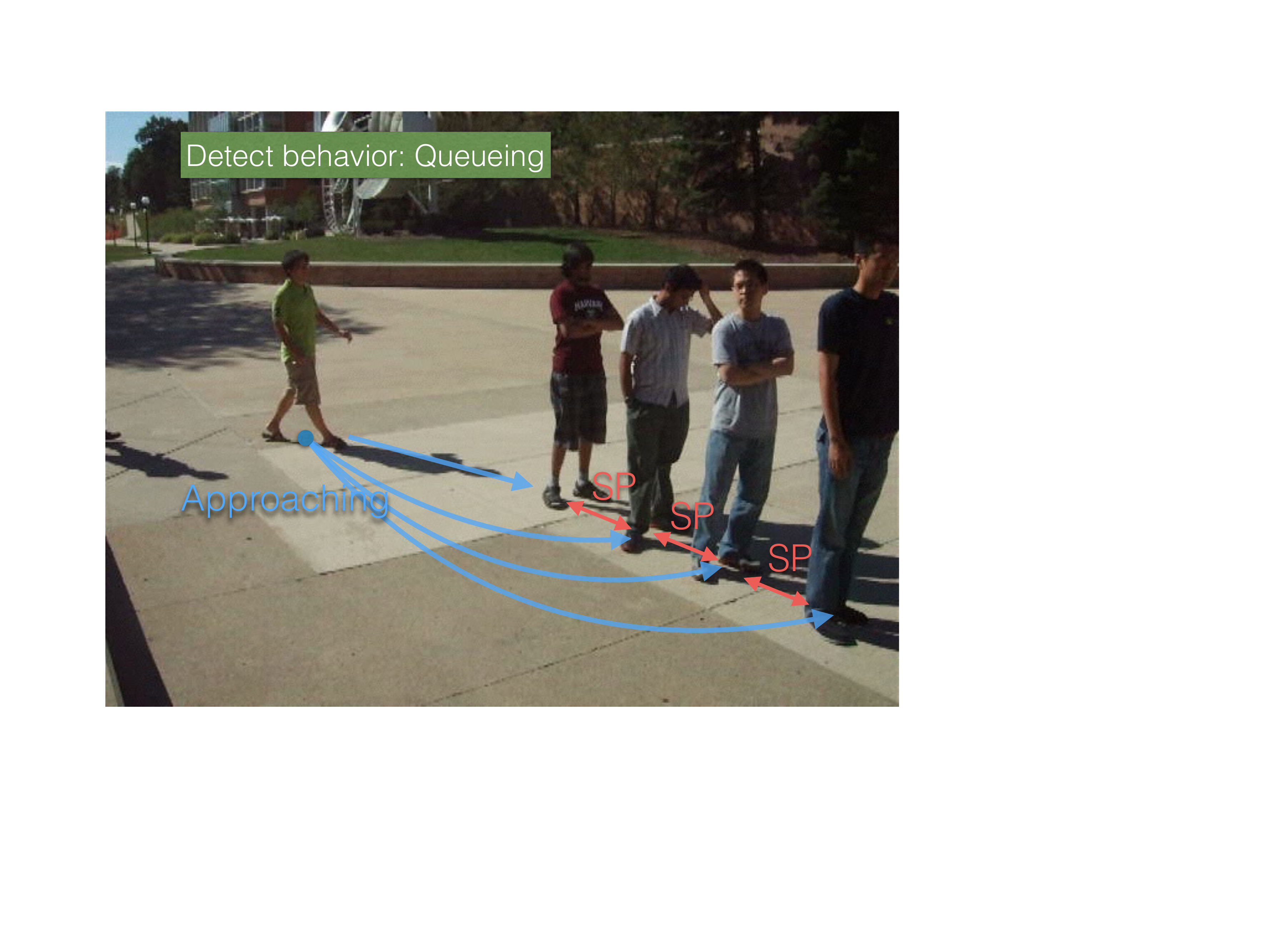}} 
  \caption{Dataset selected frames showing interactions and behavior estimates.}
  \label{fig:collective_frames}
\end{figure}

It is also important to note that, unlike other methods, the proposed method can identify the role of each pair of pedestrians involved in a given collective behavior due to the hierarchical
nature of our approach.  Fig.~\ref{fig:collective_frames:cha} illustrates some frames of different detected collective behaviors, along with the corresponding pairwise
interactions. For instance, the higher-level chasing event shown in Fig.~\ref{fig:collective_frames:cha} is characterized by \emph{following} and \emph{walking-together}
pairwise lower-level events.

Despite the high overall and per-class accuracies, the system is inherently unable to differentiate activities that depend on the subject orientation at static positions. An example of this is the inability to distinguish events of ``waiting to cross a street'' and ``queueing''. However, we feel that being able to generalize to any camera setup is far a better choice for surveillance systems than to use an image-based classifier for subject orientation, which might be crucial to detect
just a few specific events.

\subsection{Cross-Dataset Analysis}
\label{sec:cross:dataset}
To investigate the potential of our method in generalizing across different scenarios (and more importantly, different camera setups), we carried out a set of experiments by training the model using one dataset and evaluating on another one. More precisely, we used the N-CAD dataset for training and the BEHAVE dataset~\citep{Blunsden:2010} for evaluation purposes. The first 7 sequences of the first set in BEHAVE were used, yielding a total of around 63k frames. The dataset presents almost the same activities of the ones described by \citet{Choi:2012} and used in this paper, with the inclusion of \emph{fighting} events and the lack of \emph{queuing} events. Upon analysis, it was clear to us that fighting would not be possible to recognize using solely trajectories -- thus, we decide to ignore the frames in which such event appears. Additionally, some sequences presented wrong annotations or no annotation at all. We carefully annotated and corrected the interactions, bounding boxes and collective activities. We made this data publicly available\footnote{\url{https://github.com/gustavofuhr/behave_comp_anno}}  to stimulate a broader application of this dataset in the future. 

It is worth noting that the interactions in the BEHAVE dataset are mainly \emph{standing pair}, \emph{splitting}, \emph{approaching}, \emph{walking-together} with very few \emph{being-followed} and \emph{following} samples. However, our method has shown an impressive capability of recognizing interaction in new, unseen sequences. 

\begin{figure}[ht!]
  \centering
  \subfloat[Pairwise interactions]{\label{fig:behave_interactions}\includegraphics[height=.21\textwidth]{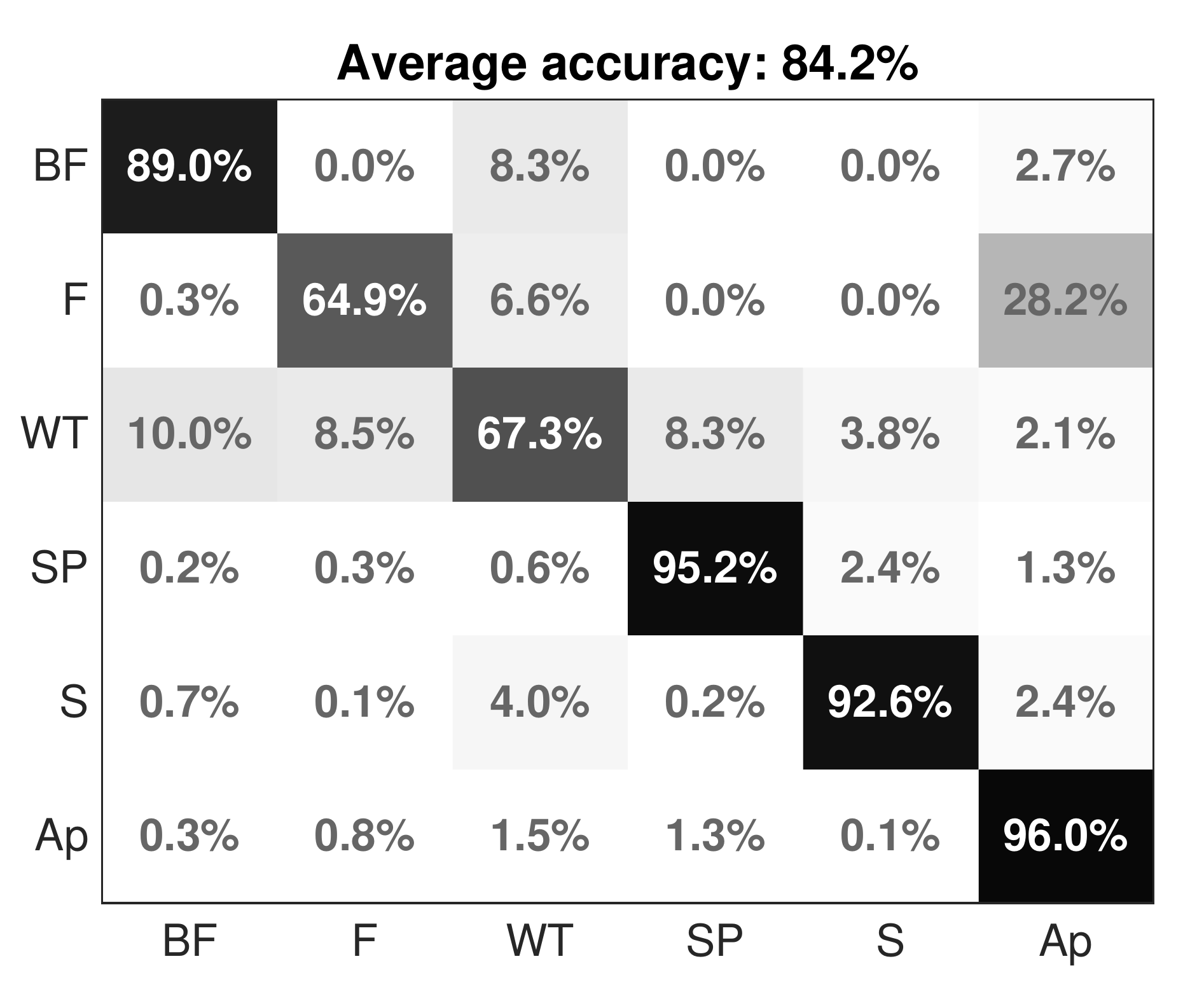}}~
\subfloat[Collective behaviors]{\label{fig:behave_collective}\includegraphics[height=.21\textwidth]{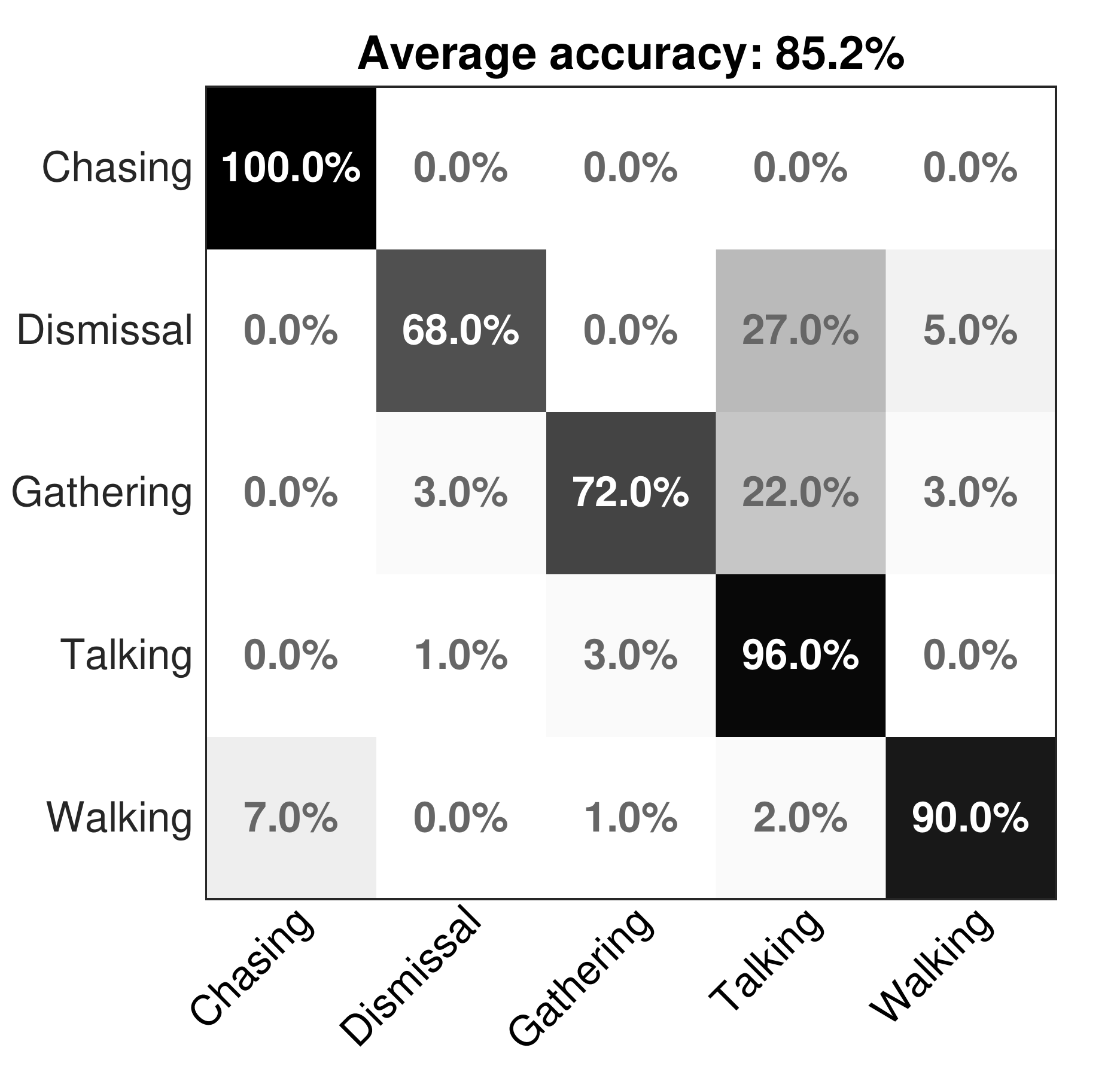}}
    \caption{Interaction and collective behavior estimates using only N-CAD as the training set and BEHAVE as testing.}
  \label{fig:behave_results}
  \vspace{-4mm}
\end{figure}

Finally, we apply the same cross-dataset validation protocol for testing collective behaviors, training with N-CAD and testing with BEHAVE. The parameters of the proposed method were kept the same, except for removing the PCA-based shape descriptor $p(t)$ defined by Equation~\eqref{eq:pca}. The reason behind this choice is that for the BEHAVE dataset there is no need to differentiate between standing groups, since there are no \emph{queuing} events. Also, we removed the three sequences from N-CAD that were labeled as \emph{Queuing}, so that this class was completely ignored in the experiment.  The results can be observed in Fig.~\ref{fig:behave_collective}, and they  indicate that our method is capable of generalizing for different camera setups in the context of collective behavior. Moreover, it does not require too much training in terms of variability to correctly estimate the collective behavior. We believe that this comes from the formulation of a two-stage approach, which helps the understanding of new instances of previously trained behaviors. One instance where this is clear comes from some of the \emph{Dismissal} events -- and analogous to \emph{Gathering}. In the BEHAVE dataset, these events usually appear as a part of a group leaving the remaining subjects of a standing group. On the other hand, in the N-CAD, the \emph{Dismissal} events always concern all subjects (no subject is left). Despite that, because we based our descriptor on interactions and shape dynamics using ground plane trajectories, we can generalize across different exemplars of the same class.

\section{Conclusions}
\label{sec:conclusions}

We proposed in this paper a novel method to describe and detect interactions and collective activities using only trajectory information in a surveillance scenario. We presented novel compact descriptors (PID and CBD) that combine cues with different natures, which are fed to a two-layered Random Forest to achieve the final classification.  

Our experimental results showed that the proposed method achieves average accuracy comparable to competitive
approaches for the N-CAD dataset, with the highest minimum per-class accuracy amongst all tested methods.
It also presented the lowest standard deviation considering the per-class accuracies, indicating a homogeneous distribution of the classification error.

The cross-dataset validation experiment also indicated that our method is able to generalize
well across different camera setups, due to the use of ground plane trajectories and very small feature vectors. To our knowledge, this is the first cross-dataset experiment reported in the context of collective behavior detection, with successful results. We believe that the lack of such experiments is due to the fact that existing methods rely directly on image-based features, which are inherently camera-dependent. Hence, training on a given camera setup and testing on a different one would probably lead to poor results.

Future work will concentrate on filtering the pool of interactions to reduce noise in the collective activity recognition, and extending the method for different, more complex activities that involve a sequence of collective behaviors, such as pick-pocketing. 

\begin{thebibliography}{26}
\expandafter\ifx\csname natexlab\endcsname\relax\def\natexlab#1{#1}\fi
\providecommand{\url}[1]{\texttt{#1}}
\providecommand{\href}[2]{#2}
\providecommand{\path}[1]{#1}
\providecommand{\DOIprefix}{doi:}
\providecommand{\ArXivprefix}{arXiv:}
\providecommand{\URLprefix}{URL: }
\providecommand{\Pubmedprefix}{pmid:}
\providecommand{\doi}[1]{\href{http://dx.doi.org/#1}{\path{#1}}}
\providecommand{\Pubmed}[1]{\href{pmid:#1}{\path{#1}}}
\providecommand{\bibinfo}[2]{#2}
\ifx\xfnm\relax \def\xfnm[#1]{\unskip,\space#1}\fi
\bibitem[{Amer et~al.(2014)Amer, Lei and Todorovic}]{Amer:2014}
\bibinfo{author}{Amer, M.R.}, \bibinfo{author}{Lei, P.},
  \bibinfo{author}{Todorovic, S.}, \bibinfo{year}{2014}.
\newblock \bibinfo{title}{Hirf: Hierarchical random field for collective
  activity recognition in videos}, in: \bibinfo{booktitle}{European Conference
  on Computer Vision}. \bibinfo{publisher}{Springer}, pp.
  \bibinfo{pages}{572--585}.
\bibitem[{Blunsden and Fisher(2010)}]{Blunsden:2010}
\bibinfo{author}{Blunsden, S.}, \bibinfo{author}{Fisher, R.},
  \bibinfo{year}{2010}.
\newblock \bibinfo{title}{The behave video dataset: ground truthed video for
  multi-person behavior classification}.
\newblock \bibinfo{journal}{Annals of the BMVA} \bibinfo{volume}{4},
  \bibinfo{pages}{4}.
\bibitem[{Breiman and Schapire(2001)}]{Breiman01randomforests}
\bibinfo{author}{Breiman, L.}, \bibinfo{author}{Schapire, E.},
  \bibinfo{year}{2001}.
\newblock \bibinfo{title}{Random forests}, in: \bibinfo{booktitle}{Machine
  Learning}, pp. \bibinfo{pages}{5--32}.
\bibitem[{Chang et~al.(2015)Chang, Zheng and Zhang}]{Chang:2015}
\bibinfo{author}{Chang, X.}, \bibinfo{author}{Zheng, W.S.},
  \bibinfo{author}{Zhang, J.}, \bibinfo{year}{2015}.
\newblock \bibinfo{title}{Learning person--person interaction in collective
  activity recognition}.
\newblock \bibinfo{journal}{IEEE Transactions on Image Processing}
  \bibinfo{volume}{24}, \bibinfo{pages}{1905--1918}.
\bibitem[{Cheng et~al.(2014)Cheng, Qin, Huang, Yan and Tian}]{Cheng:2014}
\bibinfo{author}{Cheng, Z.}, \bibinfo{author}{Qin, L.}, \bibinfo{author}{Huang,
  Q.}, \bibinfo{author}{Yan, S.}, \bibinfo{author}{Tian, Q.},
  \bibinfo{year}{2014}.
\newblock \bibinfo{title}{Recognizing human group action by layered model with
  multiple cues}.
\newblock \bibinfo{journal}{Neurocomputing} .
\bibitem[{Choi and Savarese(2012)}]{Choi:2012}
\bibinfo{author}{Choi, W.}, \bibinfo{author}{Savarese, S.},
  \bibinfo{year}{2012}.
\newblock \bibinfo{title}{A unified framework for multi-target tracking and
  collective activity recognition}, in: \bibinfo{booktitle}{European Conference
  on Computer Vision}. \bibinfo{publisher}{Springer}, pp.
  \bibinfo{pages}{215--230}.
\bibitem[{Choi and Savarese(2014)}]{Choi:2014}
\bibinfo{author}{Choi, W.}, \bibinfo{author}{Savarese, S.},
  \bibinfo{year}{2014}.
\newblock \bibinfo{title}{Understanding collective activitiesof people from
  videos}.
\newblock \bibinfo{journal}{IEEE Transactions on Pattern Analysis and Machine
  Intelligence} \bibinfo{volume}{36}, \bibinfo{pages}{1242--1257}.
\bibitem[{Choi et~al.(2009)Choi, Shahid and Savarese}]{Choi:2009}
\bibinfo{author}{Choi, W.}, \bibinfo{author}{Shahid, K.},
  \bibinfo{author}{Savarese, S.}, \bibinfo{year}{2009}.
\newblock \bibinfo{title}{What are they doing?: Collective activity
  classification using spatio-temporal relationship among people}, in:
  \bibinfo{booktitle}{International Conference on Computer Vision Workshops},
  pp. \bibinfo{pages}{1282--1289}.
\bibitem[{Feng and Bhanu(2015)}]{Feng:2015}
\bibinfo{author}{Feng, L.}, \bibinfo{author}{Bhanu, B.}, \bibinfo{year}{2015}.
\newblock \bibinfo{title}{Understanding dynamic social grouping behaviors of
  pedestrians}.
\newblock \bibinfo{journal}{IEEE Journal of Selected Topics in Signal
  Processing} \bibinfo{volume}{9}, \bibinfo{pages}{317--329}.
\bibitem[{Fern{\'a}ndez-Delgado et~al.(2014)Fern{\'a}ndez-Delgado, Cernadas,
  Barro and Amorim}]{fernandez2014we}
\bibinfo{author}{Fern{\'a}ndez-Delgado, M.}, \bibinfo{author}{Cernadas, E.},
  \bibinfo{author}{Barro, S.}, \bibinfo{author}{Amorim, D.},
  \bibinfo{year}{2014}.
\newblock \bibinfo{title}{Do we need hundreds of classifiers to solve real
  world classification problems}.
\newblock \bibinfo{journal}{J. Mach. Learn. Res} \bibinfo{volume}{15},
  \bibinfo{pages}{3133--3181}.
\bibitem[{Fuhr and Jung(2015)}]{Fuhr:2015}
\bibinfo{author}{Fuhr, G.}, \bibinfo{author}{Jung, C.}, \bibinfo{year}{2015}.
\newblock \bibinfo{title}{Camera self-calibration based on non-linear
  optimization and applications in surveillance systems}.
\newblock \bibinfo{journal}{IEEE Transactions on Circuits and Systems for Video
  Technology} .
\bibitem[{F{\"u}hr and Jung(2014)}]{fuhr2014combining}
\bibinfo{author}{F{\"u}hr, G.}, \bibinfo{author}{Jung, C.R.},
  \bibinfo{year}{2014}.
\newblock \bibinfo{title}{Combining patch matching and detection for robust
  pedestrian tracking in monocular calibrated cameras}.
\newblock \bibinfo{journal}{Pattern Recognition Letters} \bibinfo{volume}{39},
  \bibinfo{pages}{11--20}.
\bibitem[{Hall(1973)}]{Hall:1973}
\bibinfo{author}{Hall, E.T.}, \bibinfo{year}{1973}.
\newblock \bibinfo{title}{The silent language.}
\newblock \bibinfo{publisher}{Anchor}.
\bibitem[{Hwang et~al.(1994)Hwang, Lay and Lippman}]{kde_comparison_tsp94}
\bibinfo{author}{Hwang, J.}, \bibinfo{author}{Lay, S.},
  \bibinfo{author}{Lippman, A.}, \bibinfo{year}{1994}.
\newblock \bibinfo{title}{Nonparametric multivariate density estimation: a
  comparative study}.
\newblock \bibinfo{journal}{IEEE Transactions on Signal Processing}
  \bibinfo{volume}{42}, \bibinfo{pages}{2795--2810}.
\bibitem[{Jacques~Jr et~al.(2007)Jacques~Jr, Braun, Soldera, Musse and
  Jung}]{Jacques:2007}
\bibinfo{author}{Jacques~Jr, J.C.S.}, \bibinfo{author}{Braun, A.},
  \bibinfo{author}{Soldera, J.}, \bibinfo{author}{Musse, S.R.},
  \bibinfo{author}{Jung, C.R.}, \bibinfo{year}{2007}.
\newblock \bibinfo{title}{Understanding people motion in video sequences using
  voronoi diagrams}.
\newblock \bibinfo{journal}{Pattern Analysis and Applications}
  \bibinfo{volume}{10}, \bibinfo{pages}{321--332}.
\bibitem[{Kaneko et~al.(2014)Kaneko, Shimosaka, Odashima, Fukui and
  Sato}]{kaneko2014fully}
\bibinfo{author}{Kaneko, T.}, \bibinfo{author}{Shimosaka, M.},
  \bibinfo{author}{Odashima, S.}, \bibinfo{author}{Fukui, R.},
  \bibinfo{author}{Sato, T.}, \bibinfo{year}{2014}.
\newblock \bibinfo{title}{A fully connected model for consistent collective
  activity recognition in videos}.
\newblock \bibinfo{journal}{Pattern Recognition Letters} \bibinfo{volume}{43},
  \bibinfo{pages}{109--118}.
\bibitem[{Kim et~al.(2018)Kim, Lee and Lee}]{kim2017discriminative}
\bibinfo{author}{Kim, P.S.}, \bibinfo{author}{Lee, D.G.}, \bibinfo{author}{Lee,
  S.W.}, \bibinfo{year}{2018}.
\newblock \bibinfo{title}{Discriminative context learning with gated recurrent
  unit for group activity recognition}.
\newblock \bibinfo{journal}{Pattern Recognition} \bibinfo{volume}{76},
  \bibinfo{pages}{149--161}.
\bibitem[{Lathuili{\`e}re et~al.(2017)Lathuili{\`e}re, Evangelidis and
  Horaud}]{lathuiliere2017recognition}
\bibinfo{author}{Lathuili{\`e}re, S.}, \bibinfo{author}{Evangelidis, G.},
  \bibinfo{author}{Horaud, R.}, \bibinfo{year}{2017}.
\newblock \bibinfo{title}{Recognition of group activities in videos based on
  single-and two-person descriptors}, in: \bibinfo{booktitle}{Applications of
  Computer Vision (WACV), 2017 IEEE Winter Conference on},
  \bibinfo{organization}{IEEE}. pp. \bibinfo{pages}{217--225}.
\bibitem[{LaViola(2003)}]{Laviola:2003}
\bibinfo{author}{LaViola, J.}, \bibinfo{year}{2003}.
\newblock \bibinfo{title}{Double exponential smoothing: an alternative to
  kalman filter-based predictive tracking}, in: \bibinfo{booktitle}{Proceedings
  of the Workshop on Virtual Environments}, pp. \bibinfo{pages}{199--206}.
\bibitem[{Oliver et~al.(2000)Oliver, Rosario and Pentland}]{Oliver:2000}
\bibinfo{author}{Oliver, N.M.}, \bibinfo{author}{Rosario, B.},
  \bibinfo{author}{Pentland, A.P.}, \bibinfo{year}{2000}.
\newblock \bibinfo{title}{A bayesian computer vision system for modeling human
  interactions}.
\newblock \bibinfo{journal}{IEEE Transactions on Pattern Analysis and Machine
  Intelligence} \bibinfo{volume}{22}, \bibinfo{pages}{831--843}.
\bibitem[{Shao et~al.(2016)Shao, Loy and Wang}]{shao2016learning}
\bibinfo{author}{Shao, J.}, \bibinfo{author}{Loy, C.C.}, \bibinfo{author}{Wang,
  X.}, \bibinfo{year}{2016}.
\newblock \bibinfo{title}{Learning scene-independent group descriptors for
  crowd understanding}.
\newblock \bibinfo{journal}{IEEE Transactions on Circuits and Systems for Video
  Technology} .
\bibitem[{Smeulders et~al.(2013)Smeulders, Chu, Cucchiara, Calderara, Dehghan
  and Shah}]{smeulders2013visual}
\bibinfo{author}{Smeulders, A.W.}, \bibinfo{author}{Chu, D.M.},
  \bibinfo{author}{Cucchiara, R.}, \bibinfo{author}{Calderara, S.},
  \bibinfo{author}{Dehghan, A.}, \bibinfo{author}{Shah, M.},
  \bibinfo{year}{2013}.
\newblock \bibinfo{title}{Visual tracking: An experimental survey}.
\newblock \bibinfo{journal}{IEEE Transactions on Pattern Analysis \& Machine
  Intelligence} , \bibinfo{pages}{1}.
\bibitem[{Solera et~al.(2015)Solera, Calderara and
  Cucchiara}]{group:detection:crowds:pami2015}
\bibinfo{author}{Solera, F.}, \bibinfo{author}{Calderara, S.},
  \bibinfo{author}{Cucchiara, R.}, \bibinfo{year}{2015}.
\newblock \bibinfo{title}{Socially constrained structural learning for groups
  detection in crowd}.
\newblock \bibinfo{journal}{IEEE Transactions on Pattern Analysis and Machine
  Intelligence} .
\bibitem[{Sun et~al.(2016)Sun, Ai and Lao}]{sun2016localizing}
\bibinfo{author}{Sun, L.}, \bibinfo{author}{Ai, H.}, \bibinfo{author}{Lao, S.},
  \bibinfo{year}{2016}.
\newblock \bibinfo{title}{Localizing activity groups in videos}.
\newblock \bibinfo{journal}{Computer Vision and Image Understanding}
  \bibinfo{volume}{144}, \bibinfo{pages}{144--154}.
\bibitem[{Wang et~al.(2017)Wang, Ni and Yang}]{wang2017recurrent}
\bibinfo{author}{Wang, M.}, \bibinfo{author}{Ni, B.}, \bibinfo{author}{Yang,
  X.}, \bibinfo{year}{2017}.
\newblock \bibinfo{title}{Recurrent modeling of interaction context for
  collective activity recognition}, in: \bibinfo{booktitle}{The IEEE Conference
  on Computer Vision and Pattern Recognition (CVPR)}, pp.
  \bibinfo{pages}{7408--7416}.
\bibitem[{Zhou et~al.(2014)Zhou, Tang and Wang}]{Zhou2014}
\bibinfo{author}{Zhou, B.}, \bibinfo{author}{Tang, X.}, \bibinfo{author}{Wang,
  X.}, \bibinfo{year}{2014}.
\newblock \bibinfo{title}{Learning collective crowd behaviors with dynamic
  pedestrian-agents}.
\newblock \bibinfo{journal}{International Journal of Computer Vision}
  \bibinfo{volume}{111}, \bibinfo{pages}{50--68}.

\end{thebibliography}

\end{document}